\documentclass[9pt,conference]{IEEEtran}
\usepackage{cite}
\usepackage{amsmath,amssymb,amsfonts}
\usepackage{algorithm}
\usepackage{algpseudocode}
\usepackage{graphicx}
\usepackage{textcomp}
\usepackage{xcolor}
\usepackage{subfigure}
\def\BibTeX{{\rm B\kern-.05em{\sc i\kern-.025em b}\kern-.08em
    T\kern-.1667em\lower.7ex\hbox{E}\kern-.125emX}}

\begin{document}
\title{Guiding Global Placement with Reinforcement Learning}

\author{\IEEEauthorblockN{Robert Kirby}
\IEEEauthorblockA{\textit{Nvidia} \\
Santa Clara, CA, USA \\
rkirby@nvidia.com}
\and
\IEEEauthorblockN{Kolby Nottingham}
\IEEEauthorblockA{\textit{Univeristy of California Irvine} \\
Irvine, CA, USA \\
knotting@uci.edu}
\and
\IEEEauthorblockN{Rajarshi Roy}
\IEEEauthorblockA{\textit{Nvidia} \\
Santa Clara, CA, USA \\
rajarshir@nvidia.com}\\
\and
\IEEEauthorblockN{Saad Godil}
\IEEEauthorblockA{\textit{Nvidia} \\
Santa Clara, CA, USA \\
sgodil@nvidia.com}
\and
\IEEEauthorblockN{Bryan Catanzaro}
\IEEEauthorblockA{\textit{Nvidia} \\
Santa Clara, CA, USA \\
bcatanzaro@nvidia.com}}

\maketitle

\IEEEpubidadjcol

\begin{abstract}
Recent advances in GPU accelerated global and detail placement have reduced the time to solution by an order of magnitude. This advancement allows us to leverage data driven optimization (such as Reinforcement Learning) in an effort to improve the final quality of placement results. In this work we augment state-of-the-art, force-based global placement solvers with a reinforcement learning agent trained to improve the final detail placed Half Perimeter Wire Length (HPWL).

We propose novel control schemes with either global or localized control of the placement process. We then train reinforcement learning agents to use these controls to guide placement to improved solutions. In both cases, the augmented optimizer finds improved placement solutions.

Our trained agents achieve an average 1\% improvement in final detail place HPWL across a range of academic benchmarks and more than 1\% in global place HPWL on real industry designs.
\end{abstract}

\maketitle

\section{Introduction}
\label{sec:introduction}
In VLSI design high quality global placement results are highly correlated to the final quality (area, performance and power) of the physical design. As one of the first stages of the physical design flow, placement decisions affect the results of all downstream design stages. Recently DREAMPlace and ABCDPlace provide an extremely fast platform for running global and detail placement\cite{DREAMPlace,ABCDPlace}. This opens up the possibility of running large numbers of exploration placement runs to find the best possible placement solution. We leverage these advancements along with recent advances in reinforcement learning and introduce an augmented state of the art placement algorithm which learns new internal heuristics which provide higher quality final solutions.

State of the art force based academic placers use the Lagrangian relaxation technique to optimize the constrained objective function which takes into account both cell density and half-perimeter wire length (HPWL). Improvements to these algorithms (such as work in ePlace and RePlAce) frequently come as improvements to heuristic rules used during the course of the optimization \cite{ePlace,RePlAce}. In this paper we investigate whether it is possible to use reinforcement learning to find better heuristics than the ones being used today. The main contributions of this paper are:
\begin{itemize}
    \item We propose the use of large scale placement exploration accelerated by GPUs to fuel a data driven approach to placement.
    \item We propose two distinct methods of augmenting force-based global placement algorithms with reinforcement learning.
    \item We introduce a unique correlated sampling strategy for reinforcement learning algorithms acting on a two dimensional action space.
    \item We demonstrate that using these methods results in a 1\% reduction in HPWL across a range of academic and industry benchmarks.
\end{itemize}

As far as we are aware this is the first attempt to use reinforcement learning to directly control the dynamics of a state of the art global placement algorithm.

The modern VLSI design flow is an iterative process. The placement stage of a given design is run many times often with incremental changes to the underlying design. A placement engine that can learn from previous iterations and apply learned strategies to provide better quality of results to later iterations is therefore desirable. To this end we also demonstrate that our learned policies retain on average 77\% of their original benefit through hundreds of synthetic netlist edits. Given that the compute time for training our agents on a new partition is significant this ability to generalize to design changes is important.

\begin{figure}[t]
    \centering
    \includegraphics[scale=0.39]{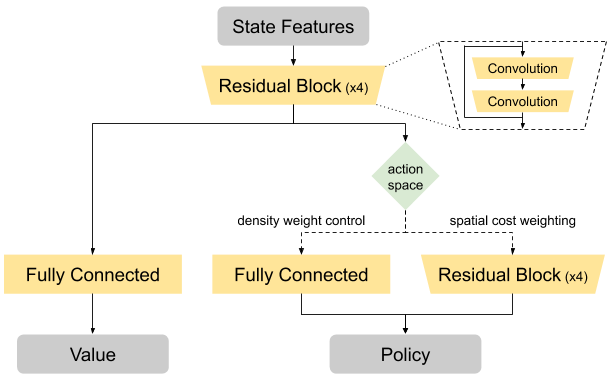}
    \caption{Neural network architecture for trained policy and value functions.}
    \label{fig:dreamnet}
\end{figure}

\section{Background}
\label{sec:background}
\subsection{Global Placement Optimization}
The objective of the global placement problem is to find locations for design cells that minimize competing objectives. These objectives always include HPWL (Half Perimeter Wire Length) and cell density, and can optionally include other metrics such as routability, timing, etc. In this work we focus only on the HPWL and cell density objectives and leave extenstion to other objectives as possible future work.

Current state of the art academic placers model the netlist and cells as an electrostatic system \cite{RePlAce,ePlace}. In these models cells are treated as point charges with the cell density cost calculated as potential energy of the system. This formulation allows for the use of the fast Fourier transform to efficiently and differentiably calculate the potential energy and therefore the density cost of a given placement. These methods then use Nesterov's method \cite{nesterov} to iteratively solve for a placement that minimizes both HPWL and density costs.

The optimization process of these approaches can be summarized as
\begin{equation} \label{eq:placement}
    \min_{\mathbf{x},\mathbf{y}} C = WL(\mathbf{x},\mathbf{y}) + \lambda D(\mathbf{x},\mathbf{y})
\end{equation}
where C is the overall cost, WL is the wire length cost function and D is the cell density cost function.

As part of the placement algorithm these solvers must choose a value ($\lambda$) to control the tradeoff between wire length and density costs (referred to as the density cost weight). Before RePlAce/ePlace, electrostatic-based solvers either chose a fixed or a gradually increasing value for the density cost weight. ePlace proposed a heuristic rule based on recent changes in wire length cost and RePlAce further suggested 'dynamic step size adaptation' which automatically adjusts the penalty based on the HPWL curve of a trial placement \cite{ePlace,RePlAce}. Using this approach they were able to show an improvement in solution quality. Additionally, the authors were able to show that adding a local density cost function which adjusted the density weight based on local overflow statistics further improved the solution quality.

Many of the insights from the previous works highlight the importance of existing heuristics to global placement tools. As the authors of previous works were able to improve results by adding heuristic rules to control the dynamics of the placement solver, in this work we study instead training a reinforcement learning agent to either replace existing heuristics or leverage new controls in order to optimize for final solution quality.

\begin{table}[tbp]
  \centering
  \caption{State Features}\label{tab:features}
    \begin{tabular}{|c|}
    \hline
    Log of current HPWL \\
    Log of $\Delta$HPWL \\
    Log of the current density weight ($\lambda)$ \\
    Current overflow as reported by DREAMPlace \\
    Current $cof$ (as defined in Algorithm \ref{lambda_scaling}) \\
    \hline
    Cell Density Map \\
    Wire Density Map \\
    Local HPWL (2x2, 4x4, 8x8, 16x16) \\
    \hline
    \end{tabular}
\end{table}

\subsection{Asynchronous Advantage Actor Critic}

We frame global placement as a Markov Decision Process (MDP) with states $\mathcal{S}$, actions $\mathcal{A}$, transition function $\mathcal{T} : \mathcal{S} \times \mathcal{A} \rightarrow \mathcal{S}$, and reward function $\mathcal{R} : \mathcal{S} \times \mathcal{A} \rightarrow \mathbb{R}$. We define $\mathcal{R}$, $\mathcal{S}$, and $\mathcal{A}$ for global placement in Sections \ref{dreamplace}, \ref{statespace}, and \ref{densityaction} or \ref{spatialcostaction} respectively. $\mathcal{T}$ is defined by DREAMPlace as explained in Section \ref{dreamplace}.

In the reinforcement learning paradigm, we optimize a  policy $\pi : \mathcal{S} \rightarrow \mathcal{A}$ to maximize expected discounted return $R=\mathbb{E}[\sum_{t=0}^T\gamma^tr_t]$ where $T$ is the horizon length, $\gamma$ is a discount factor between 0 and 1, and $r_t$ is the reward at timestep $t$ of a trajectory.

Policy gradient methods with continuous action spaces traditionally model $\pi$ as a parameterized Gaussian and an underlying function provides the parameters of this distribution.

Policy gradient methods then optimize $\pi$ directly by approximating its gradient using the objective function:

\begin{equation}
    \nabla_\theta J(\pi_\theta) = \mathbb{E}_{\tau \sim \pi_\theta} \bigg[ \sum_{t=0}^T \nabla_\theta\log\pi_\theta(a_t|s_t)A(\tau_{t:T}) \bigg]
\end{equation}

where $s_t \in \mathcal{S}$ and $a_t \in \mathcal{A}$ are the state and action at timestep $t$ of a sampled trajectory $\tau = [(s_0, a_0, r_0), ..., (s_T, a_T, r_T)]$. An agent following $\pi_\theta$ acts in an environment defined by $\mathcal{T}$ to collect $\tau$. $A(\tau)$ is the advantage function and represents how much better or worse taking one action is compared to some baseline. 

We choose the Asynchronous Advantage Actor Critic (A3C) algorithm to implement policy gradient reinforcement learning \cite{mnih2016asynchronous}. This method improves over vanilla policy gradient methods while remaining straightforward to implement and modify. A3C defines $A(\tau)$ as: 

\begin{equation}
    A(\tau) = \sum_{i=0}^{n} \gamma^i r_i+\gamma^{n+1} V_\phi(s_{n+1}) - V_\phi(s_0)
\end{equation}

where $V_\phi$ is a learned value function parameterized by $\phi$ and updated to approximate $\mathbb{E}_{\tau\sim\pi_\theta}[A(\tau)]$. This advantage function compares the return from a trajectory to the expected return from starting state $s_0$. The A3C advantage function makes use of an $n$-step Bellman target \cite{sutton1988learning}.

A3C also introduces an entropy term to the loss to aid in exploration and avoid early convergence. This is calculated with $\beta\nabla_\theta\mathbf{H}(\pi_\theta(\cdot|s_t))$ where $\mathbf{H}$ is the entropy function and $\beta$ weights the entropy term's contribution to the total loss. 

\begin{algorithm}[tbp]
\caption{$\lambda$ Scaling with Possible RL Control}\label{lambda_scaling}
\begin{algorithmic}[1]
\Procedure{$\lambda\_Scaling$}{}
    \State $\lambda \gets (\sum grad\_wl)/(\sum grad\_density)$
    \For{$k=0,last\_iteration$}
        \State $p \gets (HPWL_{k} - HPWL_{k-1})/\Delta HPWL_{ref}$
        \If{$RL Enabled$}
            \State $cof \gets RLPolicyOutput$
        \Else
            \If{$p<0$}
                \State $cof \gets cof_{max}$
            \Else
                \State $cof \gets max(cof_{min}, pow(cof_{max}, 1-p))$
            \EndIf
        \EndIf
        \State $\lambda \gets \lambda\,\times\,cof$
    \EndFor
\EndProcedure
\end{algorithmic}
\end{algorithm}

\section{Related Work}
\label{sec:relatedwork}
Some recent work has used reinforcement learning to try to solve related placement problems. \cite{mirhoseini2020chip} was able to learn a policy to explicitly place a smaller number of large macro cells before using a force based method to place the remaining cells. Other work such as \cite{FPGARL} has studied using reinforcement learning for the assignment of logic elements to FPGA logic blocks. However, these differ from our work significantly as we investigate ways to directly improve the force-based method used to place smaller standard cells. 

Many other works attempt to use machine learning to predict downstream problems during the placement stage to quickly identify potential problematic placement solutions \cite{routabilityml}\cite{routabilityml2}\cite{routingshortml}. These approaches are applications of supervised learning which provides potentially actionable information to other portions of the design flow. Our approach instead attempts to leverage reinforcement learning to learn a better placement algorithm.

The most relevant work is RePlAce/ePlace \cite{RePlAce,ePlace} and DREAMPlace \cite{DREAMPlace} which this work builds heavily upon. RePlAce and ePlace were able to push the state of the art in global placement optimization and DREAMPlace accelerated this placement using parallel processing on GPUs.

\section{Method} \label{dreamplace}

To create our reinforcement learning environment we modified DREAMPlace code to run placement in steps, yielding state information every 10 iterations and allowing the agent to observe this state and modify placement control parameters before continuing. The reward, which is provided to the agent when the density target is reached, is the percent decrease in detail place HPWL when compared to the DREAMPlace algorithm run without agent interference (the baseline HPWL). In the event the placement process diverges the environment provides a fixed reward of -10.

\subsection{DREAMPlace State} \label{statespace}

The state is presented to the policy network as a 3-dimensional tensor. The first two dimensions are spatial and the final channel dimension separates each individual feature. The features are listed in Table \ref{tab:features}. All scalar features are repeated across the first two dimensions. All features are clipped within their 10th and 90th percentile values and then normalized to zero-mean and unit-variance using the statistics of that feature during the baseline run.

\begin{table}[t]
  \centering
  \caption{Actor Critic Hyperparameters}\label{tab:hyper}
    \begin{tabular}{|c|}
    \hline
     Learning Rate ('density weight') $ = 4e-5$ \\
     Learning Rate ('spatial cost') $ = 4e-6$ \\
     $\beta = 0.05$ \\
     $n-step = 80$ \\
     Batch Size (trajectories) $ = 4$ \\
     $\gamma = 1$ \\
    \hline
    \end{tabular}
\end{table}

\begin{figure}[t]
    \centering
    \includegraphics[scale=0.3]{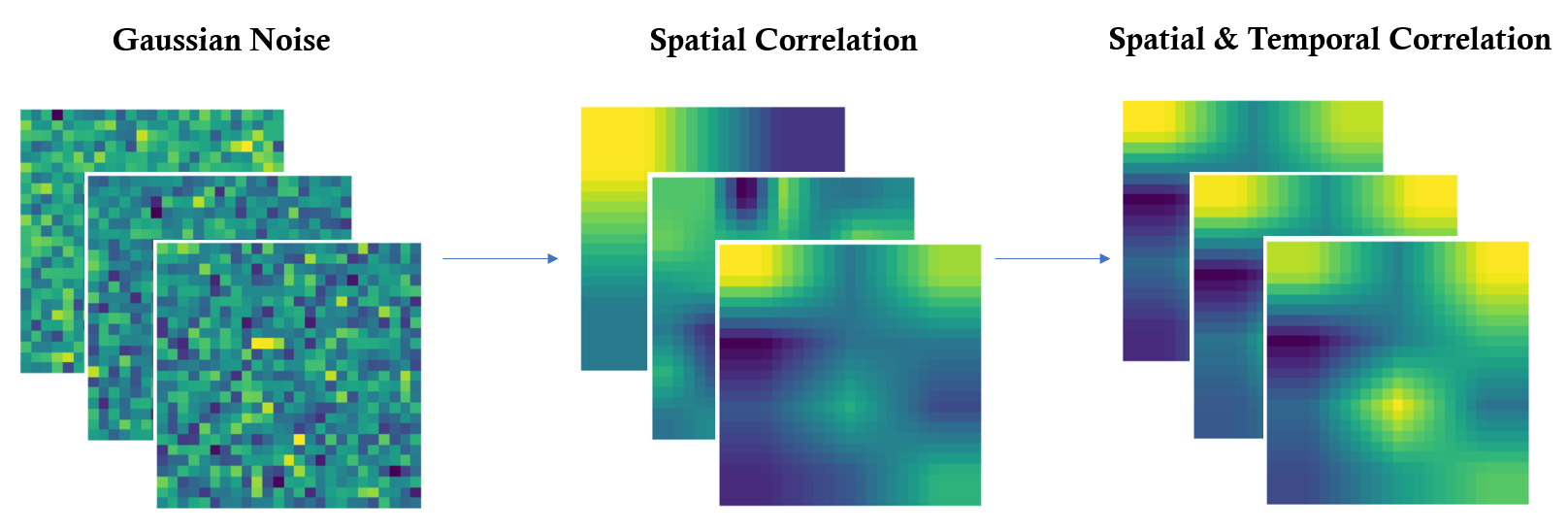}
    \caption{Visualization of spatially and temporally correlated 2D exploration noise used during training}
    \label{fig:noise}
\end{figure}

\subsection{DREAMPlace Actions}

\subsubsection{Density Weight Control} \label{densityaction}
The first of two action spaces we define is referred to as 'density weight control'. It allows the agent to set the density weight coefficient ($cof$) used to adjust the value of the density cost weight ($\lambda$) during the placement algorithm. Algorithm \ref{lambda_scaling} shows the original ePlace update rule for $\lambda$ used in DREAMPlace with an additional condition added for our RL control. When enabled, the heuristic calculation of $cof$ is instead replaced with the current output of the RL agent. This action space is therefore a single continuous value.

\subsubsection{Spatial Cost Weighting} \label{spatialcostaction}
The second action space we define is referred to as 'spatial cost weighting'. This action space allows the agent to provide a 2D field which is used to scale the gradient of the placement objective function ($C$ from Equation \ref{eq:placement}) for each cell $i$ based on the cell's location, $x_{i},y_{i}$, performing the below operation immediately after the gradient of the objective is calculated.
\begin{equation}\label{x_spatial}
    \frac{\delta C}{\delta x_{i}}_{new} =  \frac{\delta C}{\delta x_{i}} \cdot a_{t}\left[\left\lfloor\frac{x_{i}}{d}\right\rfloor,\left\lfloor\frac{y_{i}}{d}\right\rfloor\right]
\end{equation}
\begin{equation}\label{y_spatial}
    \frac{\delta C}{\delta y_{i}}_{new} =  \frac{\delta C}{\delta y_{i}} \cdot a_{t}\left[\left\lfloor\frac{x_{i}}{d}\right\rfloor,\left\lfloor\frac{y_{i}}{d}\right\rfloor\right]
\end{equation}
where $a_{t}$ is the action at the current time step and $d$ is the partition size divided by the dimension of the action space.
This approach provides the agent the ability to resist or amplify cell movement in specific areas during the placement process.

\subsection{Choice of Reinforcement Learning Algorithm}
\label{algodiscussion}
The choice of which RL algorithm to use for a given problem is an important one. Most RL algorithms fit into one of two categories, model-free or model-based, based on whether they attempt to directly model the environment dynamics. We focus on the 'model-free' category in this discussion (although we mention the possible application of 'model-based' approaches in section \ref{sec:conclusions}). Within 'model-free' RL we must decide between two styles, 'policy optimization' and 'Q learning'. In 'Q learning' an agent is trained to identify the expected future reward given a state-action pair ($Q_{\theta}(S,A)$). Alternatively 'policy optimization' methods directly train a policy function ($\pi_{\theta}(S)$) to maximize the expected future reward. 

There are a couple of reasons we chose to explore the Actor-Critic ('policy optimization') framework for this specific problem. The first is that the complete state of the placement solver is difficult to represent and therefore a significant portion of the state of the placer remains hidden from the model. This makes approximating the true $Q$ function difficult. The second is that the placement environment provides only a single reward at the terminal state. Because 'Q learning' methods train the $Q$ function to match the sum of the immediate reward and $max_{a}Q(S,a)$, it can take many samples to incorporate sparse reward information into the model. Due to the 'on policy' nature of policy optimization methods such as Actor Critic, policy model updates can instead use the final observed reward to adjust the likelihood of all actions taken during a single sampled episode.

\subsection{Actor-Critic Implementation} \label{pgalgorithm}
Our experiments use the A3C algorithm described above. We extend the asynchronous framework torchbeast with support for our multidimensional and continuous action spaces \cite{kuttler2019torchbeast}.

In our experiments we train both $\pi_\theta$ and $V_\phi$ simultaneously. We clip the gradient norm at 8 to increase training stability. To prevent the policy gradient loss from trending towards negative infinity during training we enforce a minimum value on the output of the policy standard deviation by adding a small fixed value to the network output.

Because our agent receives a sparse reward limited to terminal states, setting $\gamma=1$ assures that early actions receive equal credit and that our agent has no incentive to reach terminal states too quickly.

Additional hyper-parameters can be found in Table \ref{tab:hyper}.

\begin{algorithm}[tbp]
\caption{Netlist Edit Algorithm}\label{netlistedit}
\begin{algorithmic}[1]
\Procedure{MODIFYNETLIST}{$Netlist, NumEdits$}
    \State $n \gets 0$
    \While{$n < NumEdits$}
        \State $EditType \gets RandomEditType()$
        \If{$EditType = AddNode$}
            \State $N \gets Netlist.AddNode(RandomSize())$
            \State $P \gets $ Random integer $[2,5]$
            \For{$i \gets 0,P$}
                \State \Call{ADDNET}{$N, Netlist$}
            \EndFor
        \ElsIf{$EditType = AddNet$}
            \State $N \gets Netlist.RandomNode()$
            \State \Call{ADDNET}{$N, Netlist$}
        \ElsIf{$EditType = RemoveNode$}
            \State $N \gets Netlist.RandomNode()$
            \State $Netlist.RemoveNodeAndPins(N)$
            \State $Netlist.RemoveSinglePinNets()$
        \EndIf
        \State $n \gets n + 1$
    \EndWhile
    \State \textbf{return} InputNetlist
\EndProcedure
\Procedure{ADDNET}{$N, Netlist$}
    \State $P \gets$ Random integer $[1,4]$
    \State $NN \gets Netlist.3HopNeighborhood(N)$
    \State $Nodes \gets [N,NN.RandomNodes(P)]$
    \State $Pins \gets Netlist.AddPinsToNodes(N)$
    \State $Netlist.AddNet(Pins)$
    \State \textbf{return} $Netlist$
\EndProcedure
\end{algorithmic}
\end{algorithm}

\subsection{Neural Network Model Architectures}

In deep reinforcement learning both $\pi_\theta$ and $V_\phi$ are approximated by deep neural networks. Figure \ref{fig:dreamnet} illustrates our network architecture. In our experiments both networks share a subset of their parameters in a trunk network (a 2D convolutional neural network with residual connections). The output of the trunk network is used as input to separate value and policy branches. The value branch is composed of a single fully connected layer. In the case of our 'density weight control' action space the policy branch is also a single fully connected layer. For the 'spatial cost weighting' action space the policy branch is a fully convolutional neural network similar to the main trunk.

\begin{table*}[t]
  \centering
  \caption{Benchmark Detail Place HPWL ($\times10^{6}$) Results}\label{tab:benchmark}
    \begin{tabular}{c|cc|ccc|ccc}
    \hline
    Design & \multicolumn{2}{c|}{DREAMPlace} & \multicolumn{3}{c|}{Spatial Cost Action} &  \multicolumn{3}{c}{Density Weight Action} \\
     & \multicolumn{2}{c|}{Baseline} & HPWL & Steps & \% Improvement & HPWL & Steps & \% Improvement\\
    & HPWL & Steps & (best) & (best) & (median,best) & (best) & (best) & (median,best) \\
    \hline
    Adaptec1 & 72.78 & 63 & \textbf{72.39} & 64 & (0.51,0.53)  & 72.40 & 117 & (0.51,0.52) \\
    Adaptec2 & 81.98 & 68 & 81.42 & 73 & (0.62,0.68) & \textbf{80.85} & 172 & (1.22,1.38)  \\
    Adaptec3 & 192.75 & 70 & \textbf{190.13} & 71 & (1.25,1.35) & 191.34 & 69 & (0.62,0.73)  \\
    Adaptec4 & 173.46 & 73 & 172.66 & 76 & (0.42,0.46) & \textbf{172.58} & 151 & (0.45,0.51)  \\
    \hline
    NewBlue2 & 182.12 & 72 & \textbf{179.33} & 72 & (1.53,1.53) & 180.10 & 56 & (1.04,1.10)  \\
    NewBlue3 & 255.62 & 74 & 254.93 & 81 & (0.31,1.01) & \textbf{254.85} & 148 & (0.43,0.78)  \\
    NewBlue5 & 383.10 & 81 & 378.97 & 86 & (1.04,1.07) & \textbf{378.25} & 89 & (1.15,1.26) \\
    NewBlue6 & 441.10 & 76 & 439.26 & 81 & (0.40,0.42) & \textbf{437.75} & 121 & (0.68,0.76) \\
    NewBlue7 & 933.02 & 82 & \textbf{929.39} & 85 & (0.35,0.38) & 930.47 & 100 & (0.27,0.30)  \\
    \hline
    SuperBlue2 & 567.43 & 75 & 563.53 & 79 & (0.67,0.69) & \textbf{562.83} & 183 & (0.78,0.81)  \\
    SuperBlue3 & 289.14 & 71 & 285.63 & 83 & (1.17,1.21) & \textbf{284.61} & 204 & (1.56,1.57)  \\
    SuperBlue6 & 304.30 & 74 & 302.62 & 78 & (0.50,0.55) & \textbf{301.10} & 150 & (0.99,1.05)  \\
    SuperBlue7 & 364.26 & 76 & 361.87 & 83 & (0.53,0.65) & \textbf{359.63} & 194 & (1.22,1.27)  \\
    SuperBlue9 & 209.03 & 76 & 206.86 & 78 & (0.98,1.03) & \textbf{205.45} & 286 & (1.61,1.71) \\
    SuperBlue11 & 321.96 & 70 & 318.24 & 90 & (1.01,1.16) & \textbf{316.96} & 152 & (1.35,1.55)  \\
    SuperBlue12 & 224.10 & 103 & 221.53 & 109 & (0.96,1.15) & \textbf{220.86} & 168 & (1.41,1.44) \\
    SuperBlue14 & 210.60 & 68 & 207.84 & 80 & (1.07,1.31) & \textbf{207.27} & 180 & (1.54,1.58)  \\
    SuperBlue16 & 241.93 & 72 & 238.92 & 81 & (1.06,1.24) & \textbf{238.64} & 183 & (1.34,1.35)  \\
    SuperBlue19 & 136.57 & 68 & \textbf{133.58} & 73 & (2.14,2.18) & 134.50 & 162 & (1.48,1.52)  \\
    \hline
    Mean & - & - & - & - & (0.87,0.98) & - & - & (1.03,1.12)\\
\end{tabular}%
\end{table*}%


\begin{table*}[t]
  \centering
  \caption{Industry Global Place HPWL ($\times10^{6}$) Results}\label{tab:industry}
    \begin{tabular}{c|cc|ccc|ccc}
    \hline
    Design & \multicolumn{2}{c|}{DREAMPlace} & \multicolumn{3}{c|}{Spatial Cost Action} &  \multicolumn{3}{c}{Density Weight Action} \\
     & \multicolumn{2}{c|}{Baseline} & HPWL & Steps & \% Improvement & HPWL & Steps & \% Improvement\\
    & HPWL & Steps & (best) & (best) & (median,best) & (best) & (best) & (median,best) \\
    \hline
    Design A & 183.42 & 103 & 182.37 & 101 & (0.23,0.57) & \textbf{182.22} & 130 & (2.00,2.53) \\
    Design B & 275.61 & 90 & 274.02 & 92 & (0.30,0.58) & \textbf{268.34} & 136 & (1.29,2.64) \\
    Design C & 339.27 & 88 & 335.92 & 93 & (0.81,1.18) & \textbf{331.15} & 129 & (1.22,2.39) \\
    Design D & 259.27 & 129 & 242.83 & 135 & (5.97,6.34) & \textbf{236.66} & 196 & (4.32,8.72) \\
    Design E & 452.75 & 92 & \textbf{445.32} & 99 & (1.60,1.64) & 448.36 & 107 & (0.75,0.97) \\
    Design F & 397.11 & 103 & 393.29 & 105 & (0.86,0.96) & \textbf{391.80} & 122 & (1.23,1.34) \\
    Design G & 523.37 & 101 & \textbf{519.71} & 106 & (0.63,0.70) & 521.95 & 137 & (0.19, 0.27) \\
    Design H & 367.87 & 117 & 365.16 & 119 & (0.48,0.74) & \textbf{361.19} & 134 & (1.50, 1.82) \\
    Design I & 530.14 & 96 & \textbf{517.73} & 105 & (2.32, 2.34) & 524.30 & 116 & (0.93,1.10) \\
    \hline
    Mean & - & - & - & - & (1.47,1.67) & & - & (1.49,2.42)\\
\end{tabular}%
\end{table*}%

\begin{table*}[t]
  \centering
  \caption{Industry Dataset}\label{tab:industrydata}
    \begin{tabular}{c|cc}
    \hline
    Design & \# Standard Cells & \# Nets \\
    \hline
    Design A & 1075356 & 1004645 \\ 
    Design B & 1306336 & 1355267 \\ 
    Design C & 1344574 & 1388626 \\ 
    Design D & 1462400 & 1378091 \\ 
    Design E & 1525405 & 1528489 \\ 
    Design F & 1533034 & 1222882 \\ 
    Design G & 1616198 & 1043353 \\ 
    Design H & 1864867 & 1565958 \\ 
    Design I & 2264505 & 2276034 \\ 
    \hline
\end{tabular}%
\end{table*}%

\subsection{Correlated Noise}\label{noise}
Traditionally when drawing samples from the policy $\pi$ for a continuous action space a sample is drawn from the unit normal distribution and then scaled and shifted by the parameters provided by the policy neural network. The entropy in this parameterized distribution allows the agent to occasionally "explore" new actions even if they were not highly likely under the given policy.
For our 2D action space care must be taken to ensure 'meaningful' exploration decisions are taken. If each value in the 2D action grid is IID there is too much high frequency noise (both spatially and temporally) in the actions for the agent to chance upon exploring coherent strategies. \cite{lillicrap2019continuous,wawrzynski2015control} suggests sampling an Ornstein-Uhlenbeck process \cite{uhlenbeck1930theory} to enforce temporal consistency in the sampled noise. This process models the velocity of a Brownian particle which is both temporally correlated and mean-reverting. We extend this to also include spatial consistency across the action space. To do this for each training episode we pick a random resolution lower than the action resolution and sample each pixel from an independent Ornstein-Uhlenbeck process. We then use bilinear upsampling to interpolate this lower resolution back to the size of original action space. This forces varying amounts of spatial and temporal consistency in the exploration noise. The noise sampled from this process is visually depicted in Figure \ref{fig:noise}.
Without this correlated sampling our 'spatial cost' agent was unable to learn a policy that improved over the baseline HPWL.

\subsection{Training Setup}
Training is performed on a single DGX machine with 8 Tesla V100 GPUs. 7 GPUs are filled with actor instances (2 or 3 per GPU depending on design size) running our DREAMPlace based RL environment. The final GPU is used for running the training algorithm for the policy and value networks. Training is performed for 1.5e6 training steps (between 10,000 and 25,000 placement episodes depending on placement episode length) and the best solution found during training is reported. This training process takes between 24 and 48 hours depending on the size of the partition.

\subsection{Inference Time}
The addition of our agent to the placement process does have a small effect on runtime because we run our feature collection and agent neural network once every 10 placement steps. The addition of our agent adds a roughly 10\% overhead to the DREAMPlace global placement runtime.

\section{Experiments} \label{sec:experimentalresults}

\subsection{Data Set}
We train our reinforcement learning policy on an array of open source placement benchmarks that have been previously used for various placement competitions. These include the ISPD '05 \cite{ispd2005} and '06 \cite{ispd2006} and DAC '12 \cite{dac2012} competition. We report non-scaled detail place HPWL values from DREAMPlace and ABCDPlace to obtain the 'DREAMPlace HPWL' detail place values for all partitions. In addition we run the same experiment on 9 industry designs. Due to memory limits running the GPU accelerated detail place step, we report improvement in the global place HPWL result.

\begin{figure}[t]
    \centering
    \includegraphics[scale=0.6]{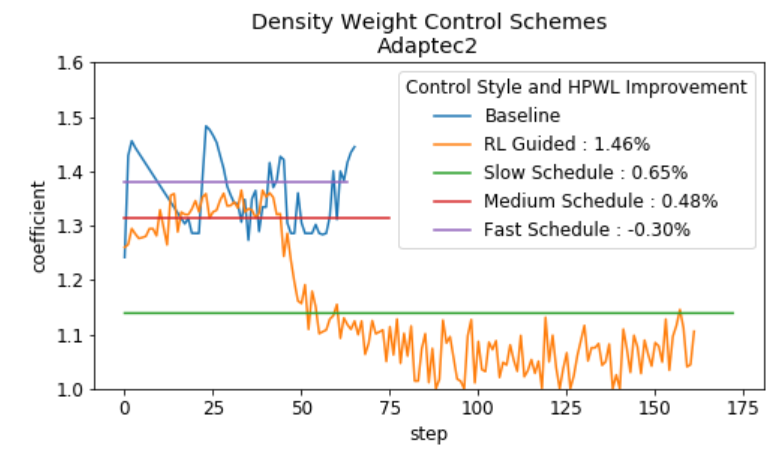}
    \caption{Various Density Weight Control Schemes and resulting HPWL Improvement: RL Guided schedule outperforms static schedules with similar numbers of iterations.}
    \label{fig:adaptec2}
\end{figure}

\begin{figure}[b]
    \centering
    \includegraphics[scale=0.2]{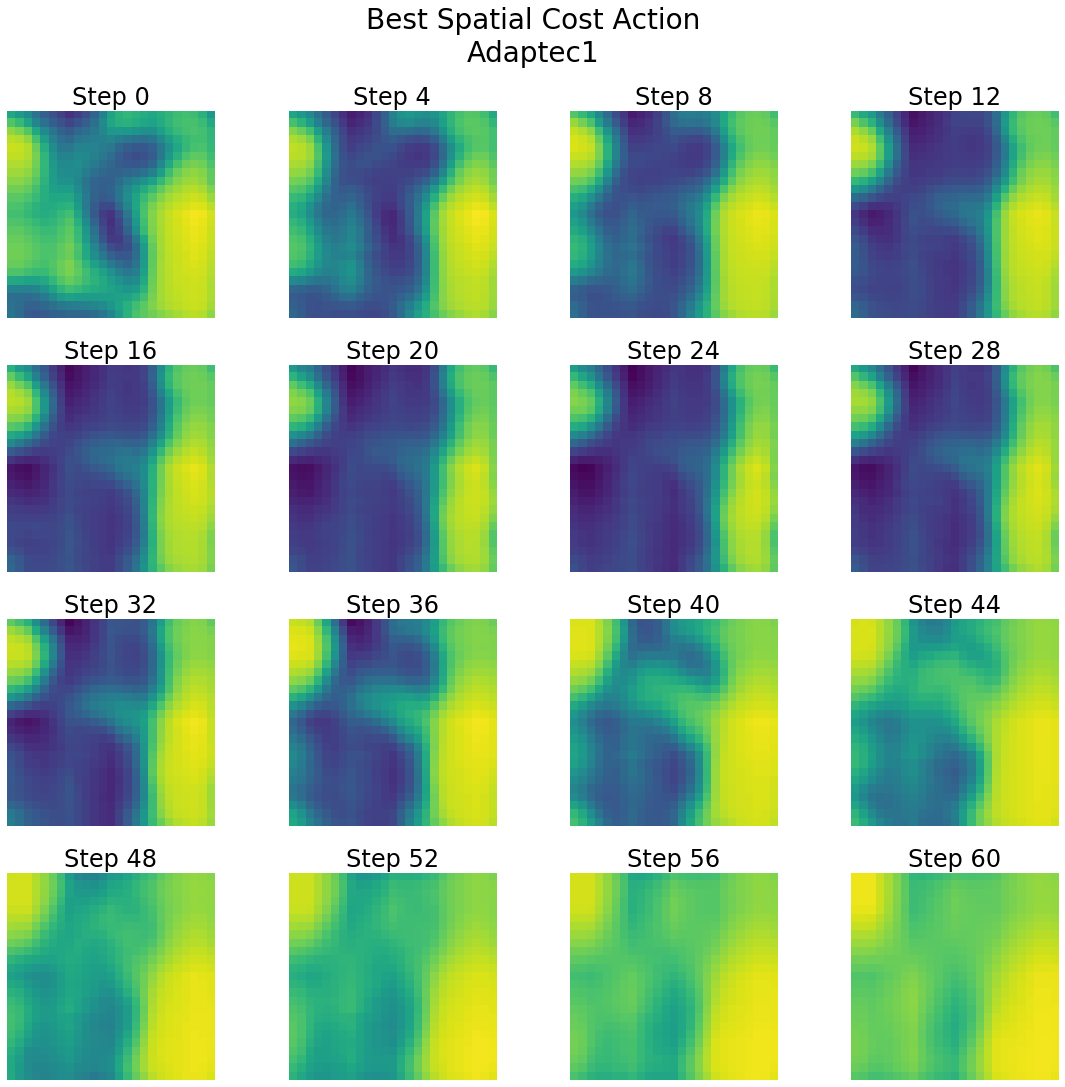}
    \caption{Adaptec1 Best Policy (spatial cost): Action $a$ from Equation \ref{x_spatial} \& \ref{y_spatial} (low-high : blue-yellow). RL agent uses spatially and temporally varying control based on locations of macros (lower right) and logic clusters (upper right) to guide placement algorithm to more optimal solution.}
    \label{fig:adaptec1}
\end{figure}

\subsection{Results}

In Table \ref{tab:benchmark} we present our results on open source benchmark designs. We report the max and median of the best placement solutions found across 3 independent training runs. The median 'density weight' action space solutions are on average 0.87\% better than the DREAMPlace baseline solutions and the 'spatial cost' action space solutions are on average 1.03\% improved over the baseline. 

In Table \ref{tab:industry} we present results in a similar manner on nine designs taken from multiple industry workflows. On these designs the agent is able to improve by 1.47\% and 1.49\% on average using the 'spatial cost' and 'density weight' actions respectively. Notably the agent is able to improve the solution for Design D by more than 5\%.
 
 Our RL guided placement methods are able to find placement solutions for all designs which have a shorter HPWL than the DREAMPlace baseline. Interestingly, one action space does not outperform the other across all benchmarks. While the 'density weight' action does appear to be superior on most designs, it comes at the cost of an increase in the number of placement iterations to convergence.

\subsection{Investigating Trained Agent Policies}

\subsubsection{Density Weight Control}

As seen in Figure \ref{fig:adaptec2}, the RePlAce rule for adjusting the density weight cost chooses a significantly different schedule than the baseline heuristic. Specifically, the agent tends to favor smaller increases in the density weight which leads to more placement iterations and longer time to converge to the target density. However, the choice of when to use small density weight updates and when to use larger ones seems to be both important and non-trivial because the final policies outperform both slow and fast static policies.

\subsubsection{Spatial Cost Weighting}

If we visualize the actions the trained 'spatial cost weighting' agent takes during placement (seen in Figure \ref{fig:adaptec1}), we observe the network leverages both the current state of placement and placement structures such as the macro positions and high density logic clusters to control the evolution of the placement process. Interestingly, unlike the 'density weight' action space the number of placement iterations to convergence is not significantly larger than the baseline placement. The agent is able to improve results with only a small increase in the number of iterations.

\begin{figure} [t]
    \centering
    \includegraphics[scale=0.6]{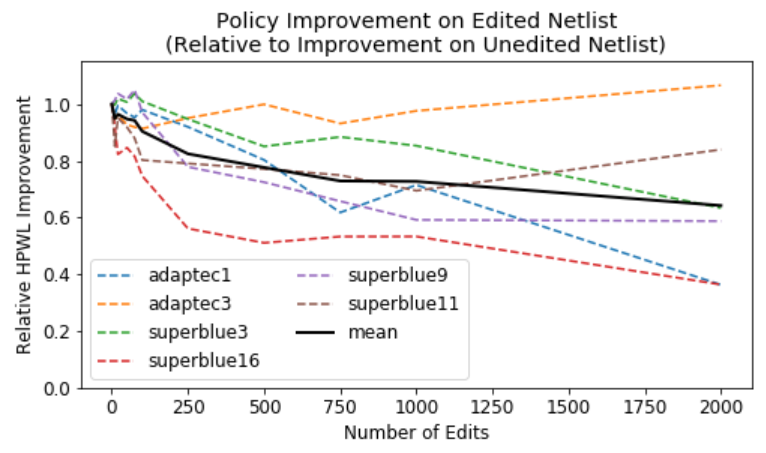}
    \caption{Effect of Netlist Edits on Trained Spatial Cost Policies: With hundreds of netlist edits, trained policies still retain around 80\% of their original improvement.}
    \label{fig:netlistedit_results}
\end{figure}

\subsection{Netlist Modification Test}
In order to investigate how robust the learned policy is to changes in the netlist we performed the following test. We make synthetic edits to the benchmark netlists and rerun DREAMPlace with our learned policy to measure how the performance of our learned policy degrades as the netlist changes. Random edits which consist of adding and removing nodes as well as adding additional nets are performed according to Algorithm \ref{netlistedit}. As shown in Figure \ref{fig:netlistedit_results}, we find that while the improvement of the learned policy degrades with these netlist edits we still see a significant improvement over the baseline placement scheme with hundreds of netlist edits. Specifically, on average we retain 78\% of the trained benefit with 500 random edits and 73\% of the benefit with 1000 random edits.

\section{Conclusion} \label{sec:conclusions}

In this work, we used GPU accelerated placement to train a reinforcement learning agent that augments a state of the art global placement algorithm. We have demonstrated this approach can learn heuristics that improve the final quality of results. Although we focus on improving half-perimeter wire length, this approach can be extended to include many other downstream metrics such as congestion, timing or power.

Ideally, an agent trained with reinforcement learning would be able to perform placement on novel designs that were unseen during training. However, we observed that the training of a generalized agent is difficult. One major challenge to progress is a lack of open source datasets. Currently, open source datasets are too small to provide sufficient coverage of the distribution of all possible designs. We invite future work to tackle this challenge of generalization. As the placement environment is deterministic, exploration of model-based approaches to improve generalization is potentially promising. Meanwhile, since designs are often built iteratively, we believe our approach can learn strategies specific to each design, yielding benefits as the design is iteratively updated.

  The result of our approach is more than 1\% reduced HPWL on both widely used academic benchmarks and industry designs. While the training time is high this benefit is still significant to highly optimized designs and demonstrates the potential of data driven approaches to global placement optimization. To our knowledge this is the first attempt to integrate reinforcement learning directly into force based global placement algorithms.


\bibliographystyle{ieeetr}
\bibliography{references}

\begin{thebibliography}{10}

\bibitem{DREAMPlace}
Y.~{Lin}, S.~{Dhar}, W.~{Li}, H.~{Ren}, B.~{Khailany}, and D.~Z. {Pan},
  ``Dreamplace: Deep learning toolkit-enabled gpu acceleration for modern vlsi
  placement,'' in {\em 2019 56th ACM/IEEE Design Automation Conference (DAC)},
  pp.~1--6, 2019.

\bibitem{ABCDPlace}
Y.~{Lin}, W.~{Li}, J.~{Gu}, H.~{Ren}, B.~{Khailany}, and D.~Z. {Pan},
  ``Abcdplace: Accelerated batch-based concurrent detailed placement on
  multi-threaded cpus and gpus,'' {\em IEEE Transactions on Computer-Aided
  Design of Integrated Circuits and Systems}, pp.~1--1, 2020.

\bibitem{ePlace}
J.~{Lu}, P.~{Chen}, C.~{Chang}, L.~{Sha}, D.~J.~H. {Huang}, C.~{Teng}, and
  C.~{Cheng}, ``eplace: Electrostatics based placement using nesterov's
  method,'' in {\em 2014 51st ACM/EDAC/IEEE Design Automation Conference
  (DAC)}, pp.~1--6, 2014.

\bibitem{RePlAce}
C.~{Cheng}, A.~B. {Kahng}, I.~{Kang}, and L.~{Wang}, ``Replace: Advancing
  solution quality and routability validation in global placement,'' {\em IEEE
  Transactions on Computer-Aided Design of Integrated Circuits and Systems},
  vol.~38, no.~9, pp.~1717--1730, 2019.

\bibitem{nesterov}
Y.~E. Nesterov, ``A method for solving the convex programming problem with
  convergence rate o$(1/k^2$),'' {\em Dokl. Akad. Nauk SSSR}, vol.~269,
  pp.~543--547, 1983.

\bibitem{mnih2016asynchronous}
V.~Mnih, A.~P. Badia, M.~Mirza, A.~Graves, T.~Lillicrap, T.~Harley, D.~Silver,
  and K.~Kavukcuoglu, ``Asynchronous methods for deep reinforcement learning,''
  in {\em International conference on machine learning}, pp.~1928--1937, 2016.

\bibitem{sutton1988learning}
R.~S. Sutton, ``Learning to predict by the methods of temporal differences,''
  {\em Machine learning}, vol.~3, no.~1, pp.~9--44, 1988.

\bibitem{mirhoseini2020chip}
A.~Mirhoseini, A.~Goldie, M.~Yazgan, J.~Jiang, E.~Songhori, S.~Wang, Y.-J. Lee,
  E.~Johnson, O.~Pathak, S.~Bae, A.~Nazi, J.~Pak, A.~Tong, K.~Srinivasa,
  W.~Hang, E.~Tuncer, A.~Babu, Q.~V. Le, J.~Laudon, R.~Ho, R.~Carpenter, and
  J.~Dean, ``Chip placement with deep reinforcement learning,'' 2020.

\bibitem{FPGARL}
R.~{Manimegalai}, E.~{Siva Soumya}, V.~{Muralidharan}, B.~{Ravindran},
  V.~{Kamakoti}, and D.~{Bhatia}, ``Placement and routing for 3d-fpgas using
  reinforcement learning and support vector machines,'' in {\em 18th
  International Conference on VLSI Design held jointly with 4th International
  Conference on Embedded Systems Design}, pp.~451--456, 2005.

\bibitem{routabilityml}
L.~{Chen}, C.~{Huang}, Y.~{Chang}, and H.~{Chen}, ``A learning-based
  methodology for routability prediction in placement,'' in {\em 2018
  International Symposium on VLSI Design, Automation and Test (VLSI-DAT)},
  pp.~1--4, 2018.

\bibitem{routabilityml2}
A.~F. {Tabrizi}, N.~K. {Darav}, L.~{Rakai}, A.~{Kennings}, and L.~{Behjat},
  ``Detailed routing violation prediction during placement using machine
  learning,'' in {\em 2017 International Symposium on VLSI Design, Automation
  and Test (VLSI-DAT)}, pp.~1--4, 2017.

\bibitem{routingshortml}
A.~F. {Tabrizi}, L.~{Rakai}, N.~K. {Darav}, I.~{Bustany}, L.~{Behjat}, S.~{Xu},
  and A.~{Kennings}, ``A machine learning framework to identify detailed
  routing short violations from a placed netlist,'' in {\em 2018 55th
  ACM/ESDA/IEEE Design Automation Conference (DAC)}, pp.~1--6, 2018.

\bibitem{kuttler2019torchbeast}
H.~Küttler, N.~Nardelli, T.~Lavril, M.~Selvatici, V.~Sivakumar,
  T.~Rocktäschel, and E.~Grefenstette, ``Torchbeast: A pytorch platform for
  distributed rl,'' 2019.

\bibitem{lillicrap2019continuous}
T.~P. Lillicrap, J.~J. Hunt, A.~Pritzel, N.~Heess, T.~Erez, Y.~Tassa,
  D.~Silver, and D.~Wierstra, ``Continuous control with deep reinforcement
  learning,'' 2019.

\bibitem{wawrzynski2015control}
P.~Wawrzynski, ``Control policy with autocorrelated noise in reinforcement
  learning for robotics,'' {\em International Journal of Machine Learning and
  Computing}, vol.~5, no.~2, p.~91, 2015.

\bibitem{uhlenbeck1930theory}
G.~E. Uhlenbeck and L.~S. Ornstein, ``On the theory of the brownian motion,''
  vol.~36, p.~823, APS, 1930.

\bibitem{ispd2005}
G.-J. Nam, C.~J. Alpert, P.~Villarrubia, B.~Winter, and M.~Yildiz, ``The
  ispd2005 placement contest and benchmarks suite,'' in {\em ISPD},
  pp.~216--220, 2005.

\bibitem{ispd2006}
G.-J. Nam, ``Ispd2006 placement contest: Benchmark suite and results,'' in {\em
  ISPD}, pp.~167--167, 2006.

\bibitem{dac2012}
N.~{Viswanathan}, C.~{Alpert}, C.~{Sze}, Z.~{Li}, and Y.~{Wei}, ``The dac 2012
  routability-driven placement contest and benchmark suite,'' in {\em DAC
  Design Automation Conference 2012}, pp.~774--782, 2012.

\end{thebibliography}

\end{document}